\documentclass[11pt]{article}

\usepackage[]{acl}

\usepackage{times}
\usepackage{latexsym}
\usepackage{booktabs}   
\usepackage{array}
\usepackage{longtable} 
\usepackage{flafter}
\usepackage[T1]{fontenc}
\usepackage{longtable}
\usepackage[utf8]{inputenc}
\usepackage{tabularx}
\usepackage{microtype}
\usepackage{booktabs}
\usepackage{graphicx}
\usepackage[dvipsnames]{xcolor}
\usepackage{inconsolata}

\usepackage{graphicx}
\usepackage{xcolor}
\usepackage[most]{tcolorbox}
\usepackage{listings}
\usepackage{xcolor}
\definecolor{gain}{RGB}{0,110,45}
\definecolor{loss}{RGB}{180,30,30}

\newcommand{\gaincell}[2]{%
  #1\,{\footnotesize\textcolor{gain}{(+#2)}}}
\newcommand{\losscell}[2]{%
  #1\,{\footnotesize\textcolor{loss}{(-#2)}}}

\newcommand{\gainbelow}[2]{%
  \begin{tabular}{@{}c@{}}
    #1\\[-1pt]
    \gaincell{}{#2}
  \end{tabular}}

\newcommand{\lossbelow}[2]{%
  \begin{tabular}{@{}c@{}}
    #1\\[-1pt]
    \losscell{}{#2}
  \end{tabular}}

\tcbuselibrary{breakable,skins}

\newtcolorbox{promptbox}[1][]{%
  breakable,
  colback=black!4,
  colframe=black!25,
  boxrule=0.4pt,
  arc=1pt,
  left=5pt,right=5pt,top=4pt,bottom=4pt,
  fontupper=\small\raggedright,
  fonttitle=\bfseries\sffamily\footnotesize,
  coltitle=black,
  colbacktitle=black!8,
  title={#1}
}

\title{Learning Perception to Action Rules through Reinforcement Learning for Theory-of-Mind Reasoning}

\author{
  Soyoung Oh\textsuperscript{1}\thanks{Equal contribution.},
  Nursulu Sagimbayeva\textsuperscript{1*},
  Nicholas Jennings\textsuperscript{2},
  Paulina Hussein\textsuperscript{1}, \\
  \textbf{Mehwish Alam\textsuperscript{3},
  and Vera Demberg\textsuperscript{1,4}} \\
  \textsuperscript{1}Saarland University, 
  \textsuperscript{2}Deutsches Forschungszentrum
  f\"ur K\"unstliche Intelligenz\\
  \textsuperscript{3}T\'el\'ecom Paris, \textsuperscript{4}Max Planck Institute for Informatics \\
  \texttt{\{soyoung,vera\}@lst.uni-saarland.de} \\
  \texttt{\{nusa00001,s8pahuss\}@stud.uni-saarland.de} \\
  \texttt{nije01@dfki.de},
  \texttt{mehwish.alam@telecom-paris.fr}
}

\begin{document}
\maketitle

\begin{abstract}

In Theory-of-Mind (ToM) reasoning, large language models may correctly infer the agent's mental state but fail to use that inference when predicting or judging the agent's subsequent actions. In this work, we investigate whether explicitly connecting perception, belief, and action with rules improves ToM reasoning. We generate concise \textit{if--then} rules that express relations between perception to action and use Group Relative Policy Optimization (GRPO) to train Qwen2.5 and Llama3 models across four ToM benchmarks. GRPO with rule generation outperforms prompting, supervised fine-tuning, and think reasoning. Moreover, rules generated by larger models ($\geq$3B) provide transferable knowledge that improves downstream performance of GPT-4o-mini. These findings suggest that explicit rules help models bridge the reasoning gap in ToM and support knowledge transfer across models.\footnote{Our code will be publicly open upon acceptance.}
\end{abstract}

\section{Introduction}

\begin{figure}[t]
    \centering
    \begin{minipage}{\linewidth}
        \centering
        \raggedright
        \small
        \textbf{Context:}
        Noor wants to make a latte for a customer who asked for oat milk. Noor grabs a milk pitcher and fills it with oat milk. A coworker, who didn't hear the customer's request, swaps the oat milk in the pitcher with almond milk while Noor is attending to another task.\par

        \smallskip
        \textbf{Action question:}
        What will Noor do?\par

        \medskip
        \includegraphics[width=\linewidth]{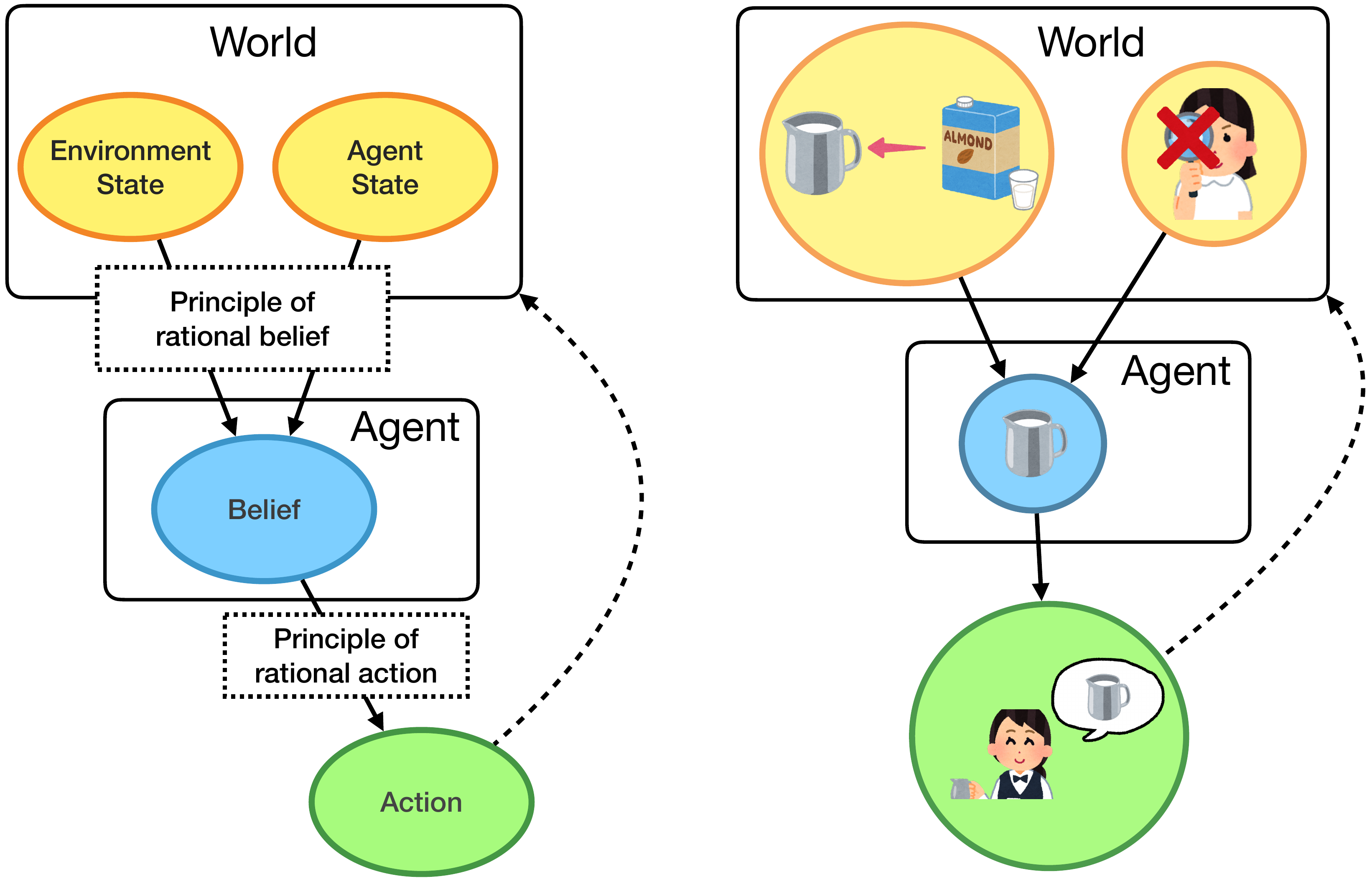}
    \end{minipage}
    \caption{Causal structure of ToM, adapted from~\citet{baker2011bayesian}. An example story and question illustrate the perception-belief-action dependencies required for reasoning about an agent's mental state.}
    \label{fig:noor-story}
\end{figure}

Humans can effortlessly infer and track others' unobservable mental states, a capacity known as Theory-of-Mind (ToM), and use these inferences to guide their social interactions~\cite{zhang2012perspective, heyes2014cultural}. Given ToM's central role in human interaction, many studies have investigated whether Large Language Models (LLMs) exhibit this ability, revealing limitations in their ToM reasoning~\cite{strachan2024testing, xu-etal-2024-opentom, gandhi2023understanding, xiao2025towards}. In particular, \citet{gu2026simpletom} showed that although LLMs can correctly answer explicit questions about agents' mental states, they struggle to predict and judge the agents' future actions, despite they already possess relevant knowledge on the agents' perceptions and beliefs. Our hypothesis is that these reasoning failures are caused by difficulties in connecting and applying relevant knowledge that models already possess, rather than an inability to make relevant inferences. This is motivated by previous findings that language models can solve individual subproblems while failing to compose their solutions into a correct final answer, which is known as the compositionality gap~\cite{press2023measuring}.

In this study, therefore, we investigate whether training LLMs with reinforcement learning (RL) to make explicit connections between inferences can bridge ToM compositionality gap. We define \emph{explicit connection} as a bridging rule that specifies how the output of one inference constrains the next inference. For example, in Figure~\ref{fig:noor-story}, adapted from~\citet{gandhi2023understanding}, correctly answering the action question requires the model to connect Noor's perception to her belief and use this information to predict that either she will replace the milk pitcher with oat milk or prepare a latte for the customer. We use LLMs to elicit story-specific connections that follow \textit{`if--then'} rule format, such as ``If Noor believes the pitcher contains the oat milk, then she will use it to prepare the latte''. We use these rules to post-train the models with Group Relative Policy Optimization (GRPO)~\cite{shao2024deepseekmath}, encouraging them to generate a rule that connects the separated inferences before answering. While prior work relies on free-form reasoning over a broad sampling space~\cite{zhong2026shortcutsreasoningrobustposttraining, lu2025theory}, we hypothesize that these concise rules constrain the reasoning space for ToM. Our main contributions are:
\begin{itemize}
    \item We introduce rule-based GRPO for ToM reasoning, which generates concise \textit{if--then} rules connecting perception, belief, and action (Section~\ref{sec:rule-gen}).

    \item Although supervising models to produce rules does not reliably improve answer prediction, GRPO with rule generation improves average accuracy by 7.2\% over free-form GRPO reasoning while using fewer thinking tokens (Section~\ref{sec:results}).


    \item Rules generated by larger models ($\geq$3B) improve GPT-4o-mini's downstream accuracy, indicating that they encode transferable task-relevant information (Section~\ref{sec:rule-util}).
    

\end{itemize}

\section{Methods}

\subsection{Data}
We conduct our experiments on four ToM benchmarks: BigToM~\cite{gandhi2023understanding}, HiToM~\cite{wu2023hi}, ToMI~\cite{le2019revisiting}, and OpenToM~\cite{xu-etal-2024-opentom}.\footnote{We report OpenToM-coarse location\_short condition.} We select these benchmarks for their focus on perception and belief tracking, with BigToM additionally testing belief–action relation, and their diversity in reasoning depth and narrative construction.



\paragraph{Data splits.}
For each benchmark, we reserve a small set of examples ($N=100$) for SFT warm-up and split the remaining examples into train, development, and test sets (Table~\ref{tab:tom-dataset-splits}). We stratify each split by the benchmark specific factor combinations (benchmark specific factors in Appendix~\ref{appx:benchmark-details}).

\begin{table}[h]
\centering
\small
\setlength{\tabcolsep}{10pt}
\renewcommand{\arraystretch}{1.}
\begin{tabular}{@{}lcc@{}}
\toprule
\textbf{Benchmark} & \textbf{Total} & \textbf{Main split} \\
\midrule
BigToM & 5,000 & 3,750 / 200 / 950 \\
HiToM & 1,200 & 720 / 200 / 180 \\
ToMi & 2,664 & 1,891 / 200 / 473 \\
OpenToM & 2,384 & 1,668 / 200 / 416 \\
\bottomrule
\end{tabular}
\caption{Dataset sizes and main splits. Main splits are reported as train/development/test.}
\label{tab:tom-dataset-splits}
\end{table}

\subsection{Rule generation}
\label{sec:rule-gen}
We generate ground truth rules using \texttt{gpt-5-mini-2025-08-07} model without revealing gold answer (prompt in Appendix~\ref{appx:prompt}). To validate the generated rules, we then prompt the same model to answer each question using its generated rule. We assume that a rule is valid if the model answers the question correctly (details in Appendix~\ref{appx:rule-val} and rule examples in Table~\ref{tab:rule-generation-examples}).

\subsection{Models}
We train and evaluate a range of Qwen2.5-Instruct-\{0.5B, 1.5B, 3B, 7B, 14B\}~\cite{yang2025qwen25m}, as well as Llama-Instruct-\{3.2-1B, 3.2-3B, 3.1-8B\} models~\cite{grattafiori2025llama}. We primarily report Qwen2.5 results in the main text and
present Llama results in Appendix~\ref{sec:llama_results}.


\subsection{Experimental setup}
We evaluate two no-training baselines. In \textbf{Base} condition, we prompt the models with story and question to predict correct answer. In \textbf{Base$_r$}, we prompt the models additionally to generate a rule before the answer (prompts in Appendix~\ref{appx:training-eval-prompts}). For consistent parsing and evaluation, we instruct all models to return outputs in JSON format. 
We then define four training conditions:


\begin{itemize}
\item \textbf{SFT$_a$:} Supervised fine-tuning to generate only the correct answer.
\item \textbf{SFT$_r$:} Supervised fine-tuning to generate a reference rule followed by the correct answer.
\item \textbf{GRPO$_t$:} RL to generate free-form reasoning and an answer within \texttt{``think''} and \texttt{``answer''} JSON fields, 
respectively~\cite{lu2025theory}.
\item \textbf{GRPO$_r$:} RL to generate a rule and an answer in JSON format within \texttt{``rule''} and \texttt{``answer''} fields, respectively. 
\end{itemize}


\subsection{Training}

We train the SFT and GRPO conditions on the same training examples for two epochs, following~\citet{lu2025theory}. We select the checkpoint with the best accuracy on a development set for evaluation on test set.\footnote{We use NVIDIA H100, A100 (80 GB), and A40 (48 GB) for the experiments.}

\paragraph{SFT} For both SFT conditions, we use a batch size of 2 and a learning rate of $5\times10^{-6}$. In SFT$_r$, we compute token-level cross-entropy loss over both. In SFT$_a$, we calculate loss on target answer.

\paragraph{Format warm-up for GRPO rules.} Following the prior work on cold start training for GRPO~\cite{chen2026specs}, we initialize each model with an SFT.
We train for up to three epochs, and do early stopping once format validity meets 95\% on a small development set (details in Appendix~\ref{appx:rule-val}).

\paragraph{GRPO optimization}
We implement GRPO with Dapo style token-loss normalization~\cite{yu2026dapo}. We use a training batch size of 8, a learning rate of $5\times10^{-6}$, and a KL regularization coefficient of $0.001$. For each input, we sample eight completions at a temperature of $0.8$
with a maximum completion length of 512 tokens. 
Each completion receives a format reward and an answer reward. A valid format requires a JSON object containing two string fields, \texttt{``rule''} and \texttt{``answer''}, where the rule should contain the word \texttt{if} followed by text. We define the format reward as
\[
r_{\mathrm{format}} =
\begin{cases}
+0.5, & \text{if the format is valid},\\
-0.5, & \text{otherwise}.
\end{cases}
\]

We evaluate the extracted answer using benchmark-specific evaluation and define the answer reward as
\[
r_{\mathrm{answer}} =
\begin{cases}
+2, & \text{if the answer is correct},\\
-2, & \text{otherwise}.
\end{cases}
\]

The total reward for each sampled completion is $r = r_{\mathrm{format}} + r_{\mathrm{answer}}$. We normalize the reward within each group of eight completions to compute the group relative advantages used for policy optimization. Therefore, the reward encourages conditional rule formatting and answer correctness.

\section{Results}
\label{sec:results}

Across training approaches, GRPO$_r$ achieves the strongest overall performance (74.2\%), outperforming GRPO$_t$ (67.0\%), SFT$_a$ (66.7\%), SFT$_r$ (61.4\%), Base$_r$ (52.4\%), and Base (45.5\%). We compare the results under rule conditioning with those from the other settings below.

\paragraph{Instructing the models to generate rules in context consistently improves performance.}
Instructing the models to generate rule improves accuracy in available comparisons (Table~\ref{tab:baseline-results}) in most setups across different temperatures \{0.2, 0.4, 0.8\}. Overall, the average accuracy increases by 6.9\%, indicating that rule conditioning provides a substantial benefit on average. 



\begin{table}[!h]
\centering
\small
\setlength{\tabcolsep}{2pt}
\renewcommand{\arraystretch}{1}
\begin{tabular}{@{}lcccc@{}}
\toprule
Model & HiToM & ToMi & BigToM & OpenToM \\
\midrule
\multicolumn{5}{l}{\textit{Base}} \\
Qwen2.5-0.5B & 1.5 & 39.1 & 51.4 & 52.6 \\
Qwen2.5-1.5B & 15.9 & 27.1 & 28.9 & 47.8 \\
Qwen2.5-3B & 7.2 & 55.3 & 73.6 & 53.6 \\
Qwen2.5-7B & 27.8 & 55.3 & 78.2 & 46.5 \\
Qwen2.5-14B & 45.0 & 62.2 & 82.1 & 59.3 \\
\midrule
\multicolumn{5}{l}{\textit{Base$_r$}} \\
\addlinespace[2pt]
Qwen2.5-0.5B & \gainbelow{6.8}{5.4} & \lossbelow{34.8}{4.3} & \gainbelow{54.1}{2.7} & \lossbelow{45.4}{7.2} \\
\addlinespace[3pt]
Qwen2.5-1.5B & \gainbelow{18.7}{2.8} & \gainbelow{42.4}{15.3} & \gainbelow{62.1}{33.2} & \gainbelow{51.9}{4.0} \\
\addlinespace[3pt]
Qwen2.5-3B & \gainbelow{24.4}{17.2} & \lossbelow{50.9}{4.3} & \gainbelow{80.8}{7.2} & \gainbelow{58.6}{5.0} \\
\addlinespace[3pt]
Qwen2.5-7B & \gainbelow{43.0}{15.2} & \gainbelow{56.0}{0.6} & \gainbelow{85.8}{7.6} & \gainbelow{62.4}{15.9} \\
\addlinespace[3pt]
Qwen2.5-14B & \gainbelow{46.3}{1.3} & \gainbelow{64.8}{2.6} & \gainbelow{86.5}{4.4} & \gainbelow{72.6}{13.3} \\
\bottomrule
\end{tabular}
\caption{Accuracy (\%) without training averaged across different temperatures. Changes below Base$_r$ are percentage-point differences from Base: \textcolor{gain}{green} indicates improvement and \textcolor{loss}{red} indicates degradation.}
\label{tab:baseline-results}
\end{table}

\paragraph{Simply teaching rule generation with SFT degrades performance.}
Unlike rule prompting, SFT to generate rule before the answer decreases average accuracy relative to answer-only fine-tuning (Table~\ref{tab:sft-indomain}). This suggests that training to just imitate the rule and output pairs does not reliably teach models to use rules for answer prediction. Rather, the models might prioritize producing the required rules over optimizing the final answer.

\begin{table}[!h]
\centering
\small
\setlength{\tabcolsep}{2pt}
\renewcommand{\arraystretch}{1}
\begin{tabular}{@{}lcccc@{}}
\toprule
Model & HiToM & ToMi & BigToM & OpenToM \\
\midrule
\multicolumn{5}{l}{\textit{SFT$_a$}} \\
Qwen2.5-0.5B & 9.4 & 71.7 & 64.3 & 65.6 \\
Qwen2.5-1.5B & 30.0 & 72.5 & 77.2 & 69.0 \\
Qwen2.5-3B & 24.4 & 73.6 & 86.4 & 81.5 \\
Qwen2.5-7B & 49.4 & 75.1 & 91.2 & 82.5 \\
Qwen2.5-14B & 46.7 & 77.2 & 93.4 & 92.1 \\
\midrule
\multicolumn{5}{l}{\textit{SFT$_r$}} \\
\addlinespace[2pt]
Qwen2.5-0.5B & \lossbelow{6.7}{2.7} & \lossbelow{50.1}{21.6} & \gainbelow{70.1}{5.8} & \lossbelow{55.3}{10.3} \\
\addlinespace[3pt]
Qwen2.5-1.5B & \lossbelow{26.7}{3.3} & \lossbelow{57.1}{15.4} & \gainbelow{80.1}{2.9} & \lossbelow{55.8}{13.2} \\
\addlinespace[3pt]
Qwen2.5-3B & \gainbelow{25.6}{1.2} & \lossbelow{61.9}{11.7} & \gainbelow{87.2}{0.8} & \lossbelow{70.2}{11.3} \\
\addlinespace[3pt]
Qwen2.5-7B & \lossbelow{43.3}{6.1} & \lossbelow{66.6}{8.5} & \lossbelow{90.3}{0.9} & \lossbelow{77.4}{5.1} \\
\addlinespace[3pt]
Qwen2.5-14B & \gainbelow{54.4}{7.7} & \lossbelow{67.2}{10.0} & \lossbelow{92.5}{0.9} & \lossbelow{89.2}{2.9} \\
\bottomrule
\end{tabular}
\caption{SFT accuracy (\%) on test set. Changes in parentheses compare SFT$_r$ with SFT$_a$ for the same model.}
\label{tab:sft-indomain}
\end{table}

\paragraph{GRPO$_r$ improves performance while using fewer tokens than GRPO$_t$.} 
GRPO$_r$ consistently improves average performance across all model sizes, increasing overall accuracy by 7.2\% compared to GRPO$_t$ across random seeds \{3, 4, 42\} (two-sided paired t-test $p<0.01$, with 95\% CI of [6.18, 8.35] pp). GRPO$_r$ also generates substantially shorter reasoning traces, using 25.2 tokens on average compared with 52.1 tokens for GRPO$_t$ (Appendix~\ref{fig:illustration-avg-word-lengths}). Overall, these results suggest that rule-based RL provides a more effective and token-efficient optimization than unconstrained reasoning. We additionally show that GRPO$_r$ without a warm-up still outperforms GRPO$_t$ (Appendix~\ref{tab:grpo-think-vs-ns}).

\begin{table}[!h]
\centering
\small
\setlength{\tabcolsep}{2pt}
\renewcommand{\arraystretch}{1}
\begin{tabular}{@{}lcccc@{}}
\toprule
Model & HiToM & ToMi & BigToM & OpenToM \\
\midrule
\multicolumn{5}{l}{\textit{GRPO$_t$}} \\
Qwen2.5-0.5B & 10.0 & 71.2 & 72.6 & 61.1 \\
Qwen2.5-1.5B & 32.2 & 71.0 & 87.6 & 52.9 \\
Qwen2.5-3B & 32.4 & 67.9 & 90.3 & 78.8 \\
Qwen2.5-7B & 46.1 & 71.5 & 92.8 & 85.9 \\
Qwen2.5-14B & 55.7 & 75.3 & 92.9 & 91.6 \\
\midrule
\multicolumn{5}{l}{\textit{GRPO$_r$}} \\
\addlinespace[2pt]
Qwen2.5-0.5B & \gainbelow{15.9}{5.9} & \gainbelow{78.3}{7.1} & \gainbelow{84.7}{12.1} & \lossbelow{53.7}{7.4} \\
\addlinespace[3pt]
Qwen2.5-1.5B & \gainbelow{48.9}{16.7} & \gainbelow{81.6}{10.6} & \gainbelow{91.5}{3.9} & \gainbelow{53.8}{0.9} \\
\addlinespace[3pt]
Qwen2.5-3B & \gainbelow{44.8}{12.4} & \gainbelow{84.3}{16.4} & \gainbelow{92.7}{2.4} & \gainbelow{90.3}{11.5} \\
\addlinespace[3pt]
Qwen2.5-7B & \gainbelow{57.4}{11.3} & \gainbelow{82.9}{11.4} & \gainbelow{93.1}{0.3} & \gainbelow{90.5}{4.6} \\
\addlinespace[3pt]
Qwen2.5-14B & \gainbelow{68.0}{12.3} & \gainbelow{85.8}{10.5} & \gainbelow{94.0}{1.1} & \gainbelow{93.4}{1.8} \\
\bottomrule
\end{tabular}
\caption{RL accuracy (\%) on the test set averaged across three different random seeds. Changes in parentheses compare GRPO$_r$ with GRPO$_t$ for the same model.}
\label{tab:rl-indomain}
\end{table}

\subsection{Rule utility beyond trace length}
\label{sec:rule-util}

We conduct two analyses to test whether the benefits of GRPO$_r$ extend beyond just producing shorter traces. First, we constrain the length of GRPO$_t$ to match the concise reasoning traces of GRPO$_r$ on HiToM because it exhibits the largest average trace length gap (GRPO$_r$ 29.5; GRPO$_t$ 71.4) (prompt in Appendix \ref{appx:grpo-think-short-prompt}). Although shorter reasoning modestly improves performance, GRPO$_r$ consistently outperforms both long (GRPO$_t$) and short (GRPO$_{t_{s}}$) free-form thinking as described in Table~\ref{tab:length-control} (token length in Figure~\ref{fig:token-length-hitom}). 

\begin{table}[!h]
\centering
\small
\setlength{\tabcolsep}{4pt}
\begin{tabular}{@{}lrrr@{}}
\toprule
Model & GRPO$_t$ & GRPO$_{t_{s}}$ & GRPO$_r$ \\
\midrule
0.5B  & 11.1 & 10.6 & \textbf{17.8} \\
1.5B  & 33.9 & 34.4 & \textbf{46.1} \\
3B    & 32.2 & 35.0 & \textbf{46.7} \\
7B    & 44.4 & 46.1 & \textbf{60.0} \\
14B   & 56.7 & 61.1 & \textbf{65.6} \\
\bottomrule
\end{tabular}
\caption{HiToM accuracy (\%) under random seed 42. GRPO$_t$ denotes the long-thinking model, GRPO$_{t_s}$ the short-thinking model, and GRPO$_r$ the rule-generation model.}
\label{tab:length-control}
\end{table}

\begin{figure}[!ht]
  \centering
  \includegraphics[width=\columnwidth]{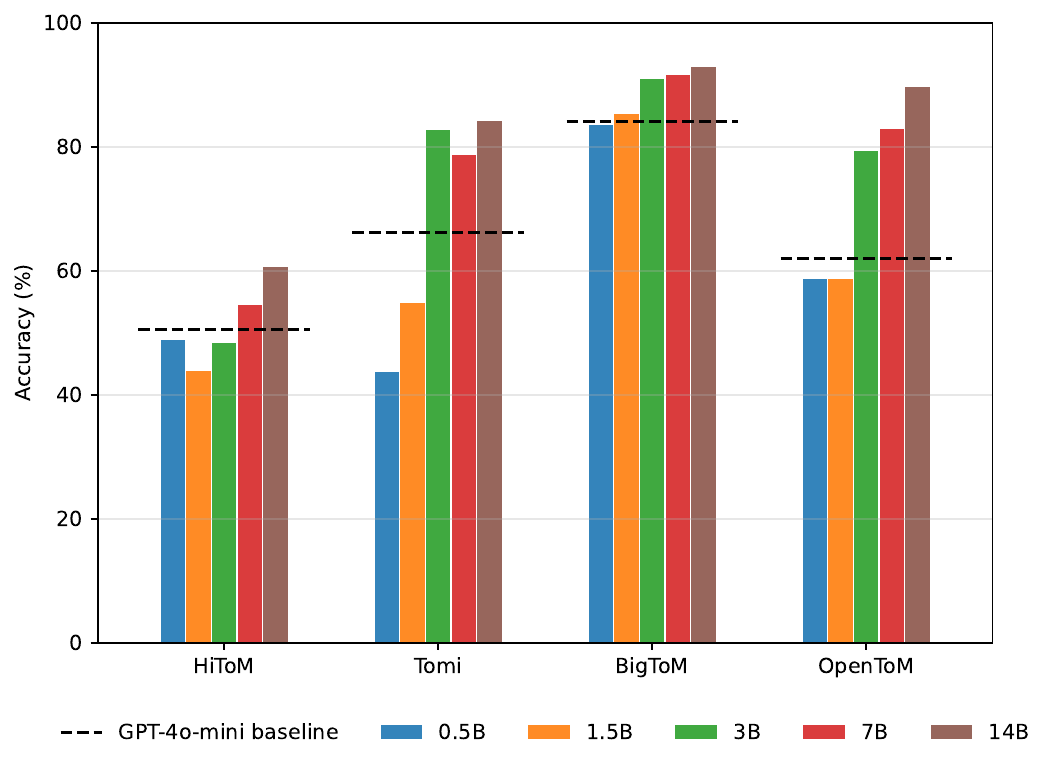}
  \caption{Downstream accuracy (\%) of GPT-4o-mini augmented with rules generated by each GRPO$_r$ models. Dashed line denotes GPT-4o-mini no-rule baseline.}
  \label{fig:rule-transfer}
\end{figure}

Second, we pass rule traces in the story context to GPT-4o-mini and measure the effect on its downstream accuracy, following~\citet{lu2025theory}. Figure~\ref{fig:rule-transfer} shows a scale-dependent transfer pattern. Rules generated by smaller models ($\leq$1.5B) reduce downstream accuracy, while the rules from larger models improve it (except 3B model for HiToM). This suggests that smaller models can produce rules that support their own generation, but lack a valid logical structure and are less transferable (qualitative analysis in Appendix~\ref{appx:error-analysis}).
In contrast, larger models tend to generate more reliable rules that can be utilized by a downstream model. 


\section{Conclusion}
We show that instructing LLMs to generate perception-belief-action rules improves ToM reasoning, whereas supervised rule imitation does not reliably improve answer prediction. GRPO with rule generation achieves the strongest accuracy with shorter traces than free-form GRPO reasoning, and rules from larger models improve downstream accuracy on GPT-4o-mini. These findings suggest that structured rule generation is effective when jointly optimized with answer correctness.




\section*{Limitations}

Our experiments cover four controlled ToM benchmarks, so the results may not generalize to benchmarks that assess ToM in more naturalistic settings. Moreover, our reward function verifies output format and answer correctness but does not directly evaluate whether a generated rule is semantically correct, or causally used for predicting the answer. Models could therefore produce uninformative rules while receiving a high reward. The observed gains may have been coming from encouraging concise intermediate traces~\cite{Renze_2024, xu2025chaindraftthinkingfaster, rakotonirina-etal-2026-correct} or from the use of structured reasoning templates, which have been shown to be particularly beneficial for smaller models~\cite{wu-etal-2026-templaterl}.

\section*{Ethical Considerations}
Our work could introduce potential risks including overreliance on models' predictions of human beliefs and intentions, and the possible use of improved social reasoning for manipulation. Generated rules may also appear credible without being semantically correct or faithfully explaining model predictions. Benchmark improvements should therefore not be interpreted as evidence of reliable mental-state understanding in real world settings.

\section*{Acknowledgements}
This project has received funding from the European Research Council (ERC) under the European Union's
Horizon 2020 research and innovation programme (ERC Starting Grant “Individualized Interaction in Discourse”,
grant agreement No. 948878). We gratefully acknowledge the stimulating research environment of the GRK 2853/1 ``Neuroexplicit Models of Language, Vision, and Action'', funded by the Deutsche Forschungsgemeinschaft (DFG, German Research Foundation) under project number 471607914. We also akcnowledge the support from a Short-Term Scientific Mission (STSM) grant provided by European Cooperation in Science and Technology (COST Action, CA23147).


\bibliography{custom}

\appendix

\section{Benchmark details}
\label{appx:benchmark-details}
Table~\ref{tab:tom-dataset-formats} summarizes the structure of five ToM benchmarks. Each dataset varies examples along multiple factors, including task condition, belief or action type, question order, and character observations or personality. The benchmarks also differ in their answer formats, BigToM, HiToM, and OpenToM
use discrete choice answers, and ToMi uses open-ended predictions. Table~\ref{tab:tom-sft-format} describes the train and development splits proportions used for a warm-up SFT before GRPO rules training.

\begin{table}[h]
\centering
\small
\setlength{\tabcolsep}{2pt}
\renewcommand{\arraystretch}{1.08}
\begin{tabular}{@{}p{1.45cm} p{3.1cm} p{2.45cm}@{}}
\toprule
\textbf{Benchmark} & \multicolumn{1}{c}{\textbf{Factors}} & \textbf{Answer format} \\
\midrule
BigToM &
condition \texttt{x} belief/action type \texttt{x} initial belief (\#25) &
Choose (a) or (b) from two options \\
\addlinespace
HiToM &
deception \texttt{x} story length \texttt{x} question order (\#30) &
One letter from 16 options \\
\addlinespace
ToMi &
false-belief order \texttt{x} question type (\#16) &
Open-ended location string \\
\addlinespace
OpenToM &
condition \texttt{x} observed \texttt{x} personality (\#12) &
Yes/No \\
\bottomrule
\end{tabular}
\caption{Benchmark stratification factors and answer formats. \#N denotes the number of possible factor combinations.}
\label{tab:tom-dataset-formats}
\end{table}

\begin{table}[t]
\centering
\small
\begin{tabular}{@{}lc@{}}
\toprule
\textbf{Dataset} & \textbf{SFT (format) warm-up} \\
\midrule
BigToM                  & 100 (75:25) \\
HiToM                   & 100 (70:30) \\
ToMi                    & 100 (83:17) \\
OpenToM & 100 (80:20) \\
\bottomrule
\end{tabular}
\caption{SFT format warm-up splits (train:development).}
\label{tab:tom-sft-format}
\end{table}

\section{Prompts}
\label{appx:prompt}

\subsection{Rule generation}
\begin{promptbox}[Rule generation]
You are an expert in Theory of Mind rules.
Your task is: given a story and a question, generate a rule that helps answer ToM question correctly.
Requirements to the rule:
\begin{itemize}
    \item Must use an IF ... THEN structure.
    \item IF and THEN should be causally related.
    \item Must be exactly one sentence.
    \item The rule can contain several conditions: e.g. IF A AND B, THEN D.
    \item The rule should contain only necessary and sufficient conditions to arrive at the decision. If some conditions are redundant, do not include them.
    \item Must be relevant to the story and its context: for example, use the names of characters and objects from the story.
    \item Must NOT simply name the answer in THEN clause: instead, capture the logic of arriving at the answer.
    \item Must be concise and not wordy.
\end{itemize}

Output only your answer in the following JSON format and nothing else:

\{

    ``rule'': "MUST be a single sentence in explicit `IF ... THEN ...' form.'',
    
\}

Input:

Story: \{\}

Question: \{\}

\textit{Answer choices: \{\} (optional)}

Rule:
\end{promptbox}





    







\begin{promptbox}[Rule validation]
Read the following story and answer the question. Apply the provided rule to arrive at the answer.

Output only your answer in the following JSON format and nothing else:

\{

"answer": "only output the answer to the question (either yes/no or a short location string, depending on the question)."

\}

Story: \{\}

Question: \{\}

Rule: \{\}

Answer:
\end{promptbox}

\subsection{Prompts for training and evaluation}
\label{appx:training-eval-prompts}

\begin{promptbox}[Base]
You are a reasoning system designed to solve Theory-of-Mind questions. You will be given a STORY and a QUESTION. Your task is to output ONLY a JSON object with an \texttt{``answer''} field:

\{\\
    ``answer'': ``The final answer''\\
\}

OUTPUT REQUIREMENTS:
\begin{itemize}
  \itemsep0.15em
  \item The \texttt{``answer''} field must contain only the final answer
        (a letter for multiple choice, or a short location string for
        open-ended questions).
  \item Output only JSON with an \texttt{``answer''} field, and no other text.
\end{itemize}
\end{promptbox}

\begin{promptbox}[Base+rule]

You are a reasoning system designed to solve Theory-of-Mind questions.
Your task is to read a STORY and a QUESTION, then produce a JSON object with two fields:

\{\\
    ``rule'': ``A rule that helps answer the question'',\\
    ``answer'': ``The final answer''\\
\}

\vspace{2pt}

OUTPUT REQUIREMENTS:

\begin{itemize}
  \itemsep0.15em
  \item The ``rule'' field should contain a rule in this form: If ..., then ....
After ``if'', put the condition(s) that are necessary to answer the question. After "then", put the conclusion that leads to the correct answer.
  \item The rule should be concise, sufficient, and helpful to answer the question correctly.
  \item The ``answer'' field MUST contain ONLY the final answer, which MUST be only the location itself (a few words, no explanation).
  
\end{itemize}
\end{promptbox}
\label{prompt:base+rule}


\begin{promptbox}[System prompt for GRPO rule]

You are a reasoning system designed to solve Theory-of-Mind questions.
Your task is to read a STORY and a QUESTION, then produce a JSON object with two fields:

\{\\
    ``rule'': "A rule that helps answer the question'',\\
    ``answer'': ``The final answer''\\
\}

OUTPUT REQUIREMENTS:
\begin{itemize}
    \item The ``rule'' field should contain a rule in this form: If ..., then ....
After ``if'', put the condition(s) that are necessary to answer the question. 
After ``then'', put the conclusion that leads to the correct answer. 
    \item The rule should be concise, sufficient, and helpful to answer the question correctly. 
    \item The ``answer'' field MUST contain ONLY the final answer (a correct letter if it is a multiple choice question, or the short location string if it is an open-ended question).
    \item Output only JSON with ``rule'' and ``answer'' fields, and no other text.
\end{itemize}
\end{promptbox}


\begin{promptbox}[System prompt for GRPO think]
You are a reasoning system designed to solve Theory-of-Mind questions. Your task is to read a STORY and a QUESTION, then produce a JSON object with two fields:

\{\\
    ``think'': ``Your reasoning in response to the question'',\\
    ``answer'': ``The final answer''
\\\}

OUTPUT REQUIREMENTS:

\begin{itemize}
    \item The ``think'' field must contain your step-by-step reasoning about the question. Use it to arrive at the correct answer in ``answer''.
    \item The ``answer'' field MUST contain ONLY the final answer (a correct letter if it is a multiple choice question, or the short location string if it is an open-ended question).
    \item Output only JSON with the ``think'' and ``answer'' fields, and no other text.
\end{itemize}
\end{promptbox}


\begin{promptbox}[System prompt for GRPO think (short traces)]
You are a reasoning system designed to solve Theory-of-Mind questions.

Your task is to read a STORY and a QUESTION, then produce a JSON object with two fields:

\{\\
    ``think'': ``Your reasoning in response to the question'',\\
    ``answer'': ``The final answer''
\\\}

OUTPUT REQUIREMENTS:

\begin{itemize}
    \item The reasoning in "think" field should be concise (ideally one sentence), sufficient, and helpful to answer the question correctly.
    \item The ``answer'' field MUST contain ONLY the final answer (a correct letter if it is a multiple choice question, or the short location string if it is an open-ended question).
    \item Output only JSON with the ``think'' and ``answer'' fields, and no other text.
\end{itemize}
\label{appx:grpo-think-short-prompt}
\end{promptbox}







\section{Validation of the generated rules}
\label{appx:rule-val}
We use medium to high reasoning for rule generation and low to medium reasoning for question answering to better match the limited capabilities of the small models we evaluate. Table~\ref{tab:tom-datasets-validity} describes the number of valid rules for each split.


\begin{table}[h]
\centering
\small
\setlength{\tabcolsep}{6pt}
\renewcommand{\arraystretch}{1.15}
\begin{tabular}{@{}lcc@{}}
\toprule
\textbf{Dataset}
  & \textbf{Training}
  & \textbf{Warm-up} \\
\midrule
BigToM
  & 3488:188:884
  & 72:23 \\
HiToM
  & 533:146:144
  & 55:23 \\
ToMi
  & 1570:164:393
  & 70:15 \\
OpenToM
  & 1668:198:416
  & 80:20 \\
\bottomrule
\end{tabular}
\caption{Number of validated rules based on train, (dev), test splits.}
\label{tab:tom-datasets-validity}
\end{table}

\section{Effect of SFT warm-up on GRPO performance}
\label{appx:warmup-effect}
To control for the effect of SFT warm-up on GRPO to rule, we compare GRPO to think (GRPO$_t$) with GRPO to rule without warm-up (GRPO$_{r_{ns}}$). As in Table~\ref{tab:grpo-think-vs-ns}, GRPO$_{r_{ns}}$ broadly matches or exceeds GRPO$_t$, improving the mean accuracy by 0.84 points across the models and benchmarks. GRPO rules with SFT warmup additionally exceeds GRPO$_{r_{ns}}$ by 5.95 points on average.

\begin{table}[h]
\centering
\small
\setlength{\tabcolsep}{3.5pt}
\renewcommand{\arraystretch}{1}
\begin{tabular}{@{}lcccc@{}}
\toprule
Model & BigToM & HiToM & ToMi & OpenToM \\
\midrule
\multicolumn{5}{l}{\textit{GRPO$_t$}} \\
Qwen2.5-0.5B  & 75.3 & 11.1 & 70.2 & 65.1 \\
Qwen2.5-1.5B  & 86.4 & 33.9 & 70.2 & 53.4 \\
Qwen2.5-3B    & 91.5 & 32.2 & 68.9 & 79.3 \\
Qwen2.5-7B    & 93.2 & 44.4 & 69.8 & 87.5 \\
Qwen2.5-14B   & 91.9 & 56.7 & 75.1 & 92.1 \\
\midrule
\multicolumn{5}{l}{\textit{GRPO$_{r_{ns}}$}} \\
Qwen2.5-0.5B
& \gainbelow{76.4}{1.1}
& \lossbelow{10.0}{1.1}
& 70.2
& \lossbelow{54.1}{11.0} \\
Qwen2.5-1.5B
& \lossbelow{86.0}{0.4}
& \gainbelow{38.9}{5.0}
& \lossbelow{62.6}{7.6}
& \gainbelow{53.8}{0.4} \\
Qwen2.5-3B
& \lossbelow{90.3}{1.2}
& \gainbelow{38.9}{6.7}
& \gainbelow{75.9}{7.0}
& \gainbelow{83.7}{4.4} \\
Qwen2.5-7B
& \lossbelow{90.7}{2.5}
& \gainbelow{45.6}{1.2}
& \gainbelow{76.5}{6.7}
& \gainbelow{88.5}{1.0} \\
Qwen2.5-14B
& \gainbelow{93.9}{2.0}
& \gainbelow{58.9}{2.2}
& \gainbelow{78.4}{3.3}
& \lossbelow{91.6}{0.5} \\
\bottomrule
\end{tabular}
\caption{Accuracy (\%) of GRPO$_t$ and GRPO$_{ns}$. Changes in parentheses compare GRPO$_{ns}$ with GRPO$_t$ for the same model.}
\label{tab:grpo-think-vs-ns}
\end{table}


\section{Reasoning trace length analysis}
\label{appx:reasoning-traces-analysis}

\begin{figure*}[ht]
    \centering
    \includegraphics[width=1.0\textwidth]{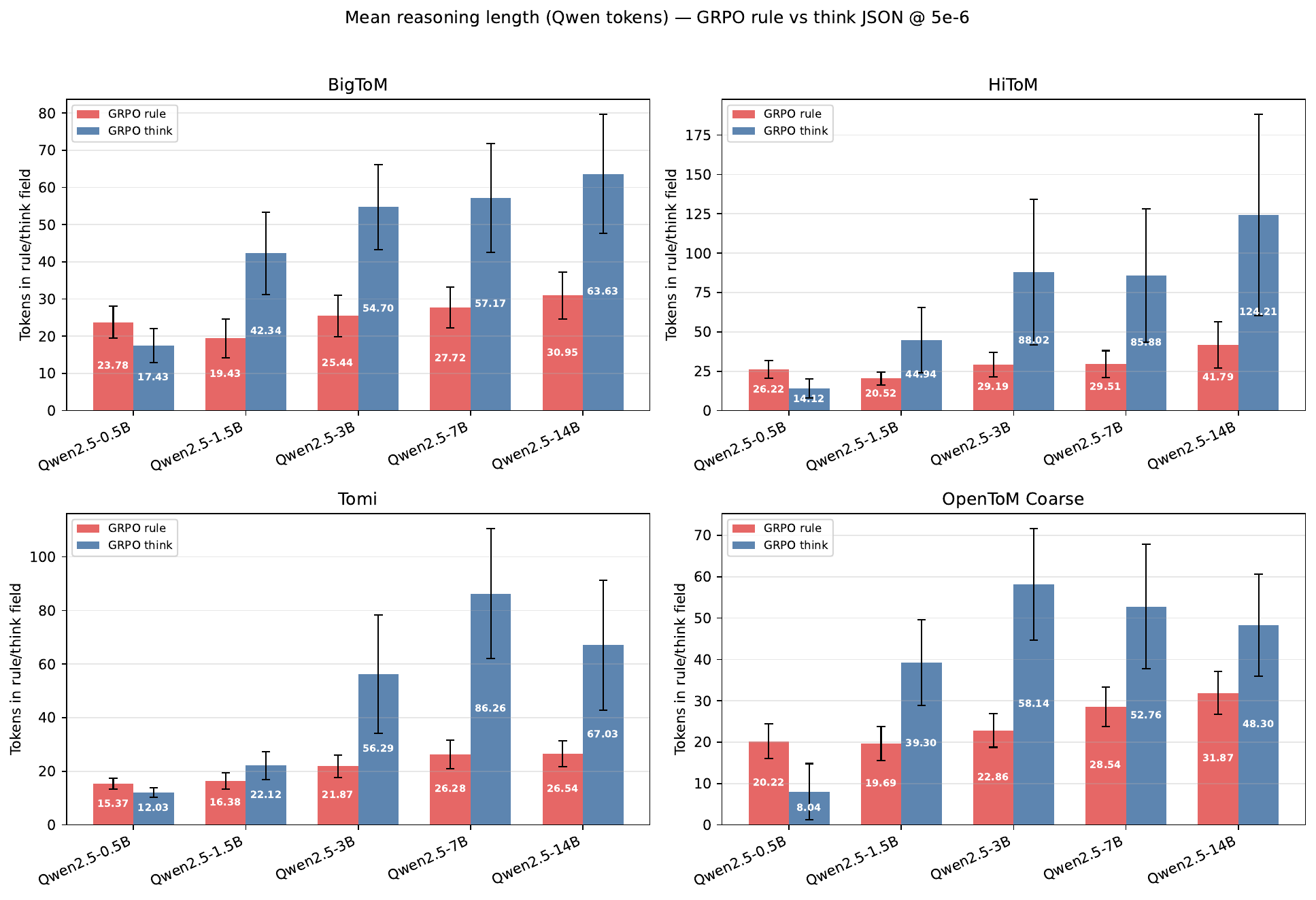}
    \caption{Average number of tokens used to produce ``rule'' vs. ``think'' reasoning}
    \label{fig:illustration-avg-word-lengths}
\end{figure*}

We first compare the average number of tokens used in \texttt{``rule''} and \texttt{``think''} traces for each model and dataset (see Figure \ref{fig:illustration-avg-word-lengths}). We find that the mean length of \texttt{``think''} reasoning is on average 1.9 times larger, with the biggest difference in Qwen 3B (\texttt{``think''} traces are on average 2.41 times longer than \texttt{``rule''} traces). 

\begin{figure*}[!h]
    \centering
    \includegraphics[width=1\textwidth]{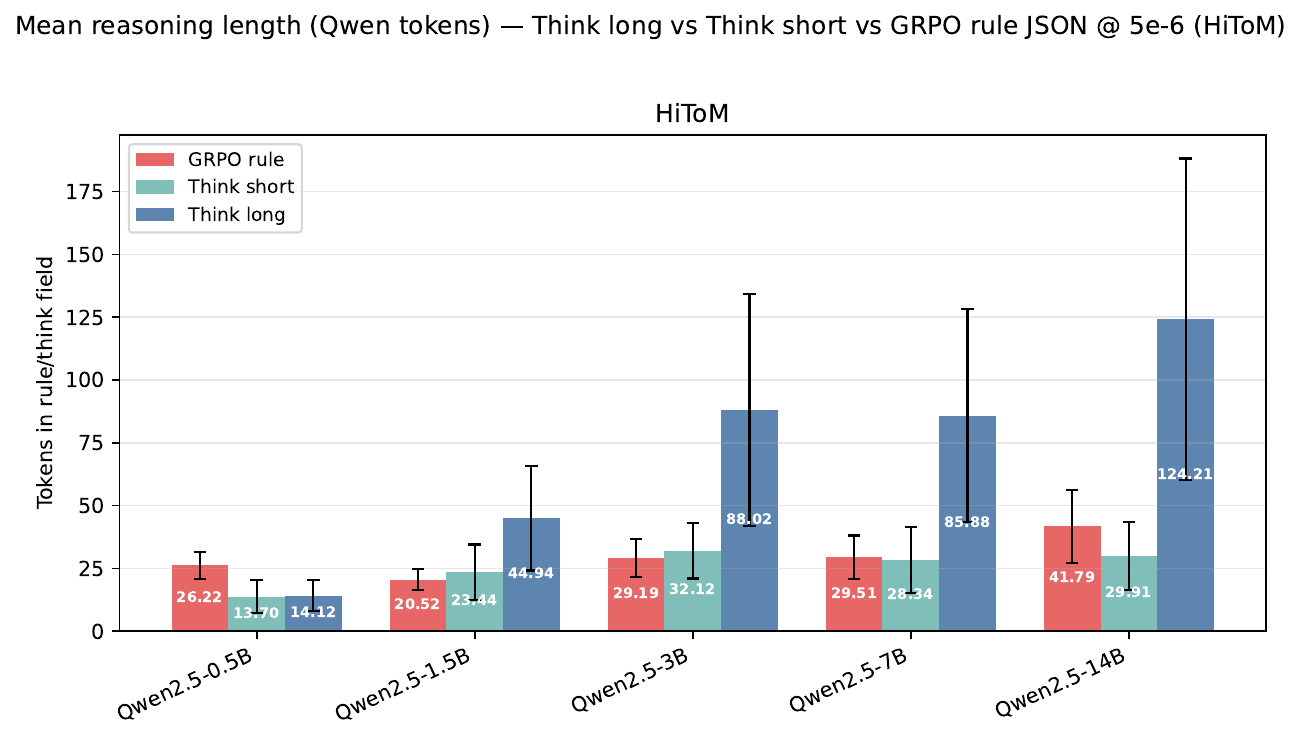}
    \caption{Average number of tokens used to produce ``rule'' vs. ``think\_short'' vs. ``think'' reasoning on HiToM benchmark}
    \label{fig:token-length-hitom}
\end{figure*}

In GRPO think\_short condition, where we instruct the models to stick to one-sentence long \texttt{``think''} traces, the average reasoning length on the HiToM benchmark reached 25.5 tokens, compared to 29.45 tokens in GRPO rule traces, and 68.61 tokens in the original GRPO think traces (Figure \ref{fig:token-length-hitom}).


\section{Qualitative analysis for generated rules from small models}
\label{appx:error-analysis}

We qualitatively analyze the rules generated by small models ($\leq$1.5B). We find that these models often produce rules that conform to the surface structure of `if-then' statements but lack semantic coherence or relevance to the question, linking unrelated story elements or asserting nonsensical relationships (examples in Table~\ref{tab:rule-generation-examples}). Models also frequently failed to incorporate the queried agent's perspective into their reasoning traces, even when their final answers were correct.

\section{Generalization to Llama family}
\label{sec:llama_results}

In order to test whether our results generalize beyond the Qwen2.5-Instruct family, we ran the same experiments on three Llama models (Llama-3.2-1B-Instruct, Llama-3.2-3B-Instruct, and Llama-3.1-8B-Instruct). We use the same training hyperparameters as for the Qwen family. The results show that GRPO$_r$ outperforms other conditions in most models and benchmarks, except for Llama-3.2-1B in OpenToM (Table~\ref{tab:llama-conditions-json-by-dataset}).

\begin{table}[!h]
\centering
\small
\setlength{\tabcolsep}{3pt}
\renewcommand{\arraystretch}{1}
\begin{tabular}{@{}lccccc@{}}
\toprule
Model & Base & SFT$_r$ & SFT$_a$ & GRPO$_r$ & GRPO$_t$ \\
\midrule
\multicolumn{6}{l}{\textit{HiToM}} \\
Llama-3.2-1B & 10.0 & 11.7 & 15.6 & \textbf{41.7} & 15.6 \\
Llama-3.2-3B & 24.4 & 30.0 & 34.4 & \textbf{51.1} & 36.7 \\
Llama-3.1-8B & 33.9 & 41.1 & 46.7 & \textbf{65.6} & 58.9 \\
\midrule
\multicolumn{6}{l}{\textit{ToMi}} \\
Llama-3.2-1B & 44.0 & 61.1 & 69.6 & \textbf{79.5} & 73.4 \\
Llama-3.2-3B & 32.6 & 63.2 & 71.7 & \textbf{81.6} & 61.7 \\
Llama-3.1-8B & 55.4 & 71.2 & 78.0 & \textbf{85.6} & 78.4 \\
\midrule
\multicolumn{6}{l}{\textit{BigToM}} \\
Llama-3.2-1B & 28.3 & 78.6 & 79.9 & \textbf{89.8} & 85.4 \\
Llama-3.2-3B & 69.3 & 85.5 & 88.0 & \textbf{92.5} & 90.4 \\
Llama-3.1-8B & 83.1 & 89.4 & 93.3 & \textbf{94.5} & 93.2 \\
\midrule
\multicolumn{6}{l}{\textit{OpenToM}} \\
Llama-3.2-1B & 47.4 & 53.6 & \textbf{64.2} & 53.6 & 53.4 \\
Llama-3.2-3B & 57.7 & 73.8 & 74.8 & \textbf{88.7} & 71.6 \\
Llama-3.1-8B & 56.7 & 85.3 & 83.9 & \textbf{93.0} & 88.9 \\
\bottomrule
\end{tabular}
\caption{Llama answer accuracy. Highlighted with \textbf{bold} is the best performing training regime for the given model and benchmark.}
\label{tab:llama-conditions-json-by-dataset}
\end{table}

\onecolumn
\footnotesize
\setlength{\tabcolsep}{4pt}
\begin{longtable}{@{}>
{\raggedright\arraybackslash}p{2.4cm} p{0.82\textwidth}@{}}
\toprule
\multicolumn{2}{@{}l}{\textbf{Dataset: HiToM}} \\
\midrule
Story &
1 Jackson, Charlotte, Avery, Liam and Benjamin entered the cellar.
2 The corn is in the blue\_treasure\_chest.
3 Jackson made no movements and stayed in the cellar for 1 minute.
4 Jackson exited the cellar.
5 Charlotte moved the corn to the green\_crate.
6 Charlotte exited the cellar.
7 Avery likes the blue\_bathtub.
8 Avery made no movements and stayed in the cellar for 1 minute.
9 Avery exited the cellar.
10 Liam made no movements and stayed in the cellar for 1 minute.
11 Liam exited the cellar.
12 Benjamin made no movements and stayed in the cellar for 1 minute.
13 Benjamin exited the cellar.
14 Jackson, Charlotte, Avery, Liam and Benjamin entered the waiting\_room. \\
Question & Where does \textbf{Jackson think Benjamin thinks} the corn is? \\
Correct answer & blue\_treasure\_chest \\
\cmidrule(lr){1-2}
Gold rule &
IF Jackson left the cellar before Charlotte moved the corn AND Benjamin witnessed Charlotte move the corn, THEN Jackson will believe that Benjamin believes the corn is at the location Benjamin saw Charlotte place it. \\[4pt]
\cmidrule(lr){1-2}

\multicolumn{2}{@{}l}{\textit{Rules}} \\
\addlinespace[2pt]
Qwen2.5-0.5B &
``\textbf{If Jackson made no movements and stayed in the cellar for 1 minute}, then Jackson thinks the corn is in the blue\_treasure\_chest.'' \\[8pt]
Qwen2.5-3B &
``If someone last saw an object in a place, they believe others also last saw it there unless they have reason to know otherwise, then the last seen place of the object is where they think others think it is.'' \\[8pt]
Qwen2.5-14B &
``If Jackson saw the corn in the blue\_treasure\_chest before exiting the cellar and did not see any movement of the corn after he left, and Benjamin never entered the cellar, then Jackson thinks Benjamin thinks the corn is in the blue\_treasure\_chest.'' \\[8pt]
\multicolumn{2}{@{}l}{\textit{Think}} \\
\addlinespace[2pt]
Qwen2.5-0.5B &
"Jackson thought Benjamin thought the corn was in the blue\_treasure\_chest." \\[4pt]
Qwen2.5-3B &
"Jackson last saw the corn in the blue\_treasure\_chest. Since he left the cellar before Benjamin did, he would not have seen Benjamin move the corn. Therefore, Jackson thinks Benjamin still thinks the corn is in the blue\_treasure\_chest." \\[4pt]
Qwen2.5-14B &
"Jackson was in the cellar when the corn was initially in the blue\_treasure\_chest. He left before Charlotte moved the corn to the green\_crate. Since Jackson did not witness the move, he would think the corn is still in the blue\_treasure\_chest. Benjamin also did not see the move as he entered after Charlotte had already moved the corn. Therefore, Benjamin would think the corn is in the green\_crate. However, since we need to determine what Jackson thinks Benjamin thinks, Jackson would assume that Benjamin thinks the corn is where Jackson last saw it, which is the blue\_treasure\_chest." \\
\midrule
\multicolumn{2}{@{}l}{\textbf{Dataset: ToMi}} \\
\midrule
Story &
1 Hannah entered the patio.
2 Noah entered the patio.
3 The sweater is in the bucket.
4 Noah exited the patio.
5 Ethan entered the study.
6 Ethan exited the study.
7 Hannah moved the sweater to the pantry. \\[4pt]
Question & Where will Noah look for the sweater? \\[4pt]
Correct answer & bucket \\
\cmidrule(lr){1-2}
Gold rule &
IF Noah was present when the sweater was in the bucket AND Noah left before Hannah moved the sweater to the pantry THEN Noah will look in the location where he last observed the sweater. \\[4pt]
\cmidrule(lr){1-2}
\multicolumn{2}{@{}l}{\textit{Rules}} \\
\addlinespace[2pt]
Qwen2.5-0.5B &
``\textbf{If Noah enters the patio}, he will enter the bucket.'' \\[4pt]
Qwen2.5-3B &
``If Noah entered before Hannah moved the sweater, then Noah will look in the bucket.'' \\[4pt]
Qwen2.5-14B &
``If Noah exits before Hannah moves the sweater, then Noah will look for the sweater where it was last known to be.'' \\[4pt]

\multicolumn{2}{@{}l}{\textit{Think}} \\
\addlinespace[2pt]
Qwen2.5-0.5B &
"Noah will look for the sweater in the bucket." \\[4pt]
Qwen2.5-3B &
"Noah last entered the patio before he exited. Since the sweater was moved from the bucket to the pantry by Hannah after Noah left, Noah would not know about this move. Therefore, Noah would still look for the sweater in the bucket." \\[4pt]
Qwen2.5-14B &
"Noah was present in the patio when the sweater was initially in the bucket (lines 2 and 3). After that, Noah left the patio (line 4), and Hannah moved the sweater from the bucket to the pantry (line 7). Since Noah did not witness the sweater being moved to the pantry, he would likely still think it is in the bucket where he last saw it." \\
\midrule
\multicolumn{2}{@{}l}{\textbf{Dataset: BigToM}} \\
\midrule
Story &
Tanya is a software engineer working on a new mobile application. She wants to ensure that the application runs smoothly and is free of bugs before launching it to the public. Tanya observes that the app's code appears to be well-structured and free of errors. Tanya believes the app's code is well-structured and free of errors. A coworker brings in donuts for the office. Tanya notices the donuts. Tanya proceeds with testing the application. \\
Question & Does Tanya believe the app's code is well-structured and free of errors or corrupted due to a server malfunction? Choose one:
(a) Tanya believes the app's code is corrupted due to a server malfunction.
(b) Tanya believes the app's code is well-structured and free of errors. \\
Correct answer & (b) Tanya believes the app's code is well-structured and free of errors. \\
\cmidrule(lr){1-2}
Gold rule &
IF Tanya, based on her observation, regards the app's code as well-structured and free of errors, THEN she will treat the code as functioning correctly and not as compromised by a server malfunction. \\[4pt]
\cmidrule(lr){1-2}
\multicolumn{2}{@{}l}{\textit{Rules}} \\
\addlinespace[2pt]
Qwen2.5-0.5B &
``If Tanya \textbf{notices the donuts,} then Tanya believes the app's code is well-structured and free of errors.'' \\[2pt]
Qwen2.5-3B &
``If Tanya observes the app's code and believes it is well-structured and free of errors, then Tanya believes the app's code is well-structured and free of errors.'' \\[8pt]
Qwen2.5-14B &
``If Tanya observes that the app's code appears to be well-structured and free of errors and she believes the app's code is well-structured and free of errors, then Tanya believes the app's code is well-structured and free of errors.'' \\[8pt]
\multicolumn{2}{@{}l}{\textit{Think}} \\
\addlinespace[2pt]
Qwen2.5-0.5B &
"Tanya observes that the app's code appears to be well-structured and free of errors." \\[4pt]
Qwen2.5-3B &
"Tanya initially observed that the app's code appears to be well-structured and free of errors. This belief is stated directly in the story: 'Tanya believes the app's code is well-structured and free of errors.' Therefore, Tanya does not believe the app's code is corrupted due to a server malfunction." \\[4pt]
Qwen2.5-14B &
"From the story, we know that Tanya observes the app's code to be well-structured and free of errors, and she believes this as well. There is no mention of any server malfunction affecting the code. Therefore, based on the information provided, Tanya believes the app's code is well-structured and free of errors." \\
\midrule
\multicolumn{2}{@{}l}{\textbf{Dataset: OpenToM}} \\
\midrule
Story &
Simon and Anderson were roommates with very different tastes. Simon couldn't stand the sight of pears, while Anderson had a fondness for their juicy sweetness.

One day, both Simon and Anderson found themselves in the bedroom at the same time. Their eyes were drawn to the pantry, where a lone pear sat, seemingly forgotten. Anderson left the room momentarily, only to return a few moments later. To his surprise, the pear was no longer in the pantry.

Simon, despite his aversion to pears, was a considerate person. Knowing that Anderson enjoyed them, he had taken it upon himself to move the pear to the dining table. He hoped that by placing it in a more visible and tempting location, his roommate would be enticed to savor the fruit. Anderson, witnessing this kind gesture, couldn't help but feel touched by Simon's selflessness.
 \\
Question & \textbf{From Anderson's perspective,} is the pear in its initial location by the end of the story? \\
Correct answer & No \\
\cmidrule(lr){1-2}
Gold rule &
IF Anderson saw the pear in the pantry, left the room, and upon returning did not see the pear in the pantry, THEN from Anderson's perspective the pear is no longer in its initial pantry location. \\
\cmidrule(lr){1-2}
\multicolumn{2}{@{}l}{\textit{Rules}} \\
\addlinespace[2pt]
Qwen2.5-0.5B &
"If Simon moved the pear to the dining table, \textbf{then Anderson felt touched by Simon's selflessness." }\\[2pt]
Qwen2.5-3B &
"If Anderson returned to the room after Simon moved the pear, then the \textbf{pear was not in its initial location.}" \\[4pt]
Qwen2.5-14B &
"If Simon moved the pear from the pantry to the dining table, then the \textbf{pear is not in its initial location by the end of the story.}" \\[4pt]
\multicolumn{2}{@{}l}{\textit{Think}} \\
\addlinespace[2pt]
Qwen2.5-0.5B &
"\textbf{Anderson} moved the pear to a more visible and tempting location." \\[4pt]
Qwen2.5-3B &
"Anderson returned to the room after a few moments and found the pear not in the pantry. Since Simon moved the pear to the dining table, and Anderson did not move it back, we can conclude that the \textbf{pear is not in its initial location.}" \\[4pt]
Qwen2.5-14B &
"Anderson initially saw the pear in the pantry. When he returned, the pear was gone. Simon moved the pear to the dining table out of consideration for Anderson. Therefore, from Anderson's perspective, the pear is not in its initial location by the end of the story." \\
\addlinespace[2pt]
\bottomrule
\caption{Examples of rule generation across model scales on ToMi, Hi-ToM, BigToM, and OpenToM. Gold rules are generated by \texttt{gpt-5-mini-2025-08-07}. In \textit{Rules} and \textit{Think} traces, highlighted with \textbf{bold} are examples of erroneous reasoning. All selected models answered correctly on the presented examples.}
\label{tab:rule-generation-examples}
\end{longtable}

\end{document}